\documentclass[10pt,twocolumn,letterpaper]{article}

\usepackage{iccv}
\usepackage{times}
\usepackage{epsfig}
\usepackage{graphicx}
\usepackage{amsmath}
\usepackage{amssymb}

\usepackage{url}
\usepackage{graphics}
\usepackage{color}
\usepackage{mathtools}
\usepackage{multirow}
\usepackage{subfigure}
\usepackage{booktabs}
\usepackage{tabularx}
\usepackage[normalem]{ulem}
\useunder{\uline}{\ul}{}
\usepackage{wrapfig,lipsum,booktabs}
\usepackage{xcolor}
\usepackage{floatflt}
\usepackage{caption}

\usepackage[pagebackref=true,breaklinks=true,letterpaper=true,colorlinks,bookmarks=false,citecolor=cyan]{hyperref}
\usepackage[ruled, lined, linesnumbered, commentsnumbered, longend]{algorithm2e}

\usepackage{caption}
\usepackage{enumitem}
\setitemize{noitemsep,topsep=0pt,parsep=0pt,partopsep=0pt}
\usepackage{flushend}

\iccvfinalcopy 

\def\iccvPaperID{****} 

\ificcvfinal\fi

\begin{document}

\title{Intrinsic-Extrinsic Preserved GANs for Unsupervised 3D Pose Transfer}

\author{Haoyu Chen$^1$ \quad
    Hao Tang$^2$ \quad 
    Henglin Shi$^1$ \quad 
    Wei Peng$^1$ \quad
    Nicu Sebe$^3$ \quad 
    Guoying Zhao$^{1,}$\thanks{Corresponding Author.}
    \\
    $^1$CMVS, University of Oulu \quad 
    $^2$Computer Vision Lab, ETH Zurich \quad 
    $^3$DISI, University of Trento
    \\
    {\tt\small \{chen.haoyu, henglin.shi, wei.peng, guoying.zhao\}@oulu.fi} \\
    {\tt\small hao.tang@vision.ee.ethz.ch \quad nicu.sebe@unitn.it}
}

\maketitle
\ificcvfinal\thispagestyle{empty}\fi

\begin{abstract}
With the strength of deep generative models, 3D pose transfer regains intensive research interests in recent years. Existing methods mainly rely on a variety of constraints to achieve the pose transfer over 3D meshes, e.g., the need for manually encoding for shape and pose disentanglement. In this paper, we present an unsupervised approach to conduct the pose transfer between any arbitrate given 3D meshes. Specifically, a novel Intrinsic-Extrinsic Preserved Generative Adversarial Network (IEP-GAN) is presented for both intrinsic (i.e., shape) and extrinsic (i.e., pose) information preservation. Extrinsically, we propose a co-occurrence discriminator to capture the structural/pose invariance from distinct Laplacians of the mesh. Meanwhile, intrinsically, a local intrinsic-preserved loss is introduced to preserve the geodesic priors while avoiding heavy computations. At last, we show the possibility of using IEP-GAN to manipulate 3D human meshes in various ways, including pose transfer, identity swapping and pose interpolation with latent code vector arithmetic. The extensive experiments on various 3D datasets of humans, animals and hands qualitatively and quantitatively demonstrate the generality of our approach. Our proposed model produces better results and is substantially more efficient compared to recent state-of-the-art methods. Code is available: \url{https://github.com/mikecheninoulu/Unsupervised_IEPGAN}
\end{abstract}
\vspace{-1em}
\section{Introduction}

\begin{figure}[!t] \small
    \centering
    \includegraphics[width=1.0\linewidth]{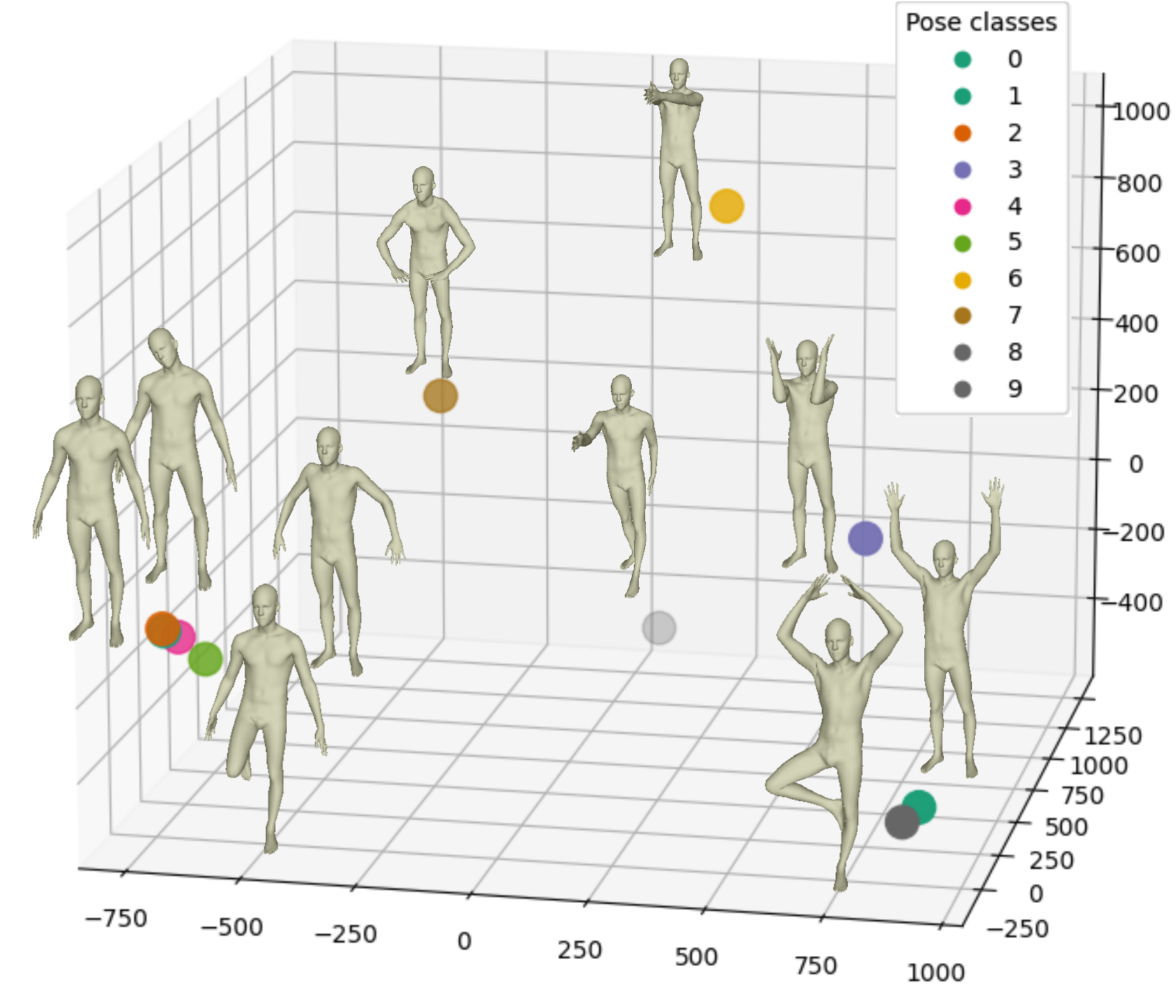}
    \caption{Visualized latent pose space distribution learned by our IEP-GAN. Dots in each color stands for a latent code of a pose class from FAUST dataset \cite{FAUST}. We observe the clusters in the latent space fit their pose representations, suggesting that the IEP-GAN has generalized the cohesive ability of projection.}
    \label{fig:latent}
    \vspace{-0.45cm}
\end{figure}

Efficient 3D mesh manipulation with high-fidelity generative models is crucial in the computer graphics field and enjoys great potential in various practical applications, ranging from 3D human activity understanding, 3D augmented reality to robotics. In this work, the focus is made to transfer the 3D pose style from the source mesh to the target mesh in an unsupervised manner. Although a few attempts have been proposed for this task, due to several issues, limitations still exist in the state-of-the-art methods.

To date, strict constraints of re-enforced correspondences of meshes are inevitable in almost all the existing methods~\cite{animal, 3dcode}. However, to provide those correspondences needs either extra manual efforts or specific requirements for data. Although some existing works claimed that they can achieve 3D pose deformation in an unsupervised setting, a constraint on the training datasets is still needed that different poses performed by the same subject should be given to successfully disentangle the shape and pose information~\cite{LIMP,Unsupervised}. This constraint actually serves as a strong supervision with manually labeled prior. Thus a real unsupervised setting for pose transfer learning is not achieved in any existing work yet. Besides, compared to 2D images that have fixed rigid permutation \cite{tang2019cycle,zhang2021controllable,tang2020bipartite,tang2020xinggan}, 3D meshes are embedded in continuous space with arbitrating orders and complex geometric attributes, which makes the task more challenging. This innately structural difference from 2D images precludes the standard discrete convolution operator to be immediately applicable to 3D meshes/points. A model specifically designed for geometric characters is desirable~\cite{Laplace}. Lastly, intrinsic (i.e., geodesic) distances are powerful metrics for learning latent representations of deformable 3D shapes~\cite{AMASS,DFAUST,FAUST}, but conventional geodesic-based methods suffer from the intensive computations, which makes them unsuitable for learning on large-scale datasets.

\emph{Learning unsupervised 3D pose transfer without any constraints needed on training sets}, this target drives our current research efforts. To this end, we propose a GAN-based framework motivated by three key observations. First, a variety of deep generative models are proposed to encode and regenerate pose-transferred meshes by disentangling pose and shape information. However, extra constraints on the training data are inevitable in all the existing methods, otherwise the shape and pose latent representations cannot be successfully disentangled, which leads to degenerate solutions. In an ideal case, a model should be able to learn the shape and pose representations without any enforced constraint on training data, see Fig.~\ref{fig:latent}. Second, to obtain the geodesic distance priors for the intrinsic preservation of meshes, various differentiable intrinsic metrics are adopted, but they either need intensive computational costs or large-scale training sets~\cite{fastgeodesic,NPT}. At last, to our knowledge, there is no existing deep generative model proposed specifically for extrinsic information learning on 3D data.

\begin{figure*}[!t] \small
    \centering
    \includegraphics[width=0.95\linewidth]{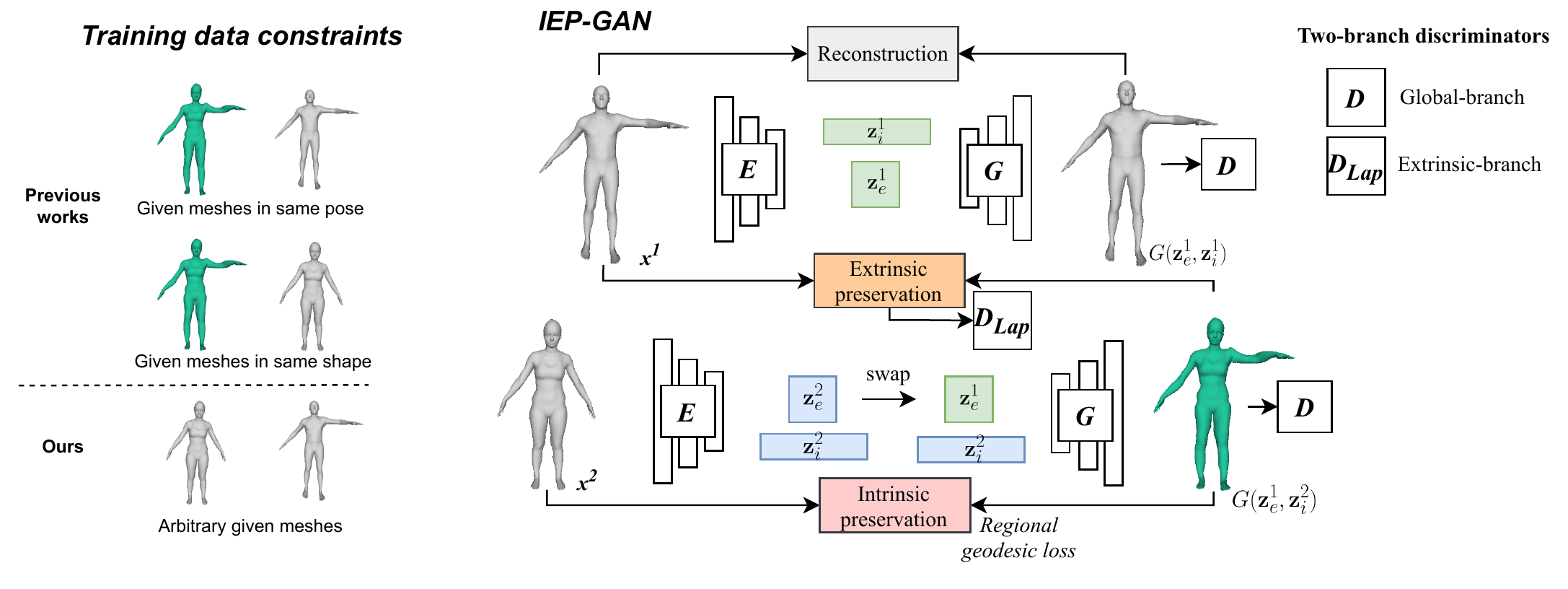}
    \caption{(\textbf{Left}) Training data constraints of previous methods and ours. Our work doesn't need any constraints on the training data while distinct requirements are inevitable in previous works to disentangle shape and pose. (\textbf{Right}) IEP-GAN framework. Our IEP-GAN consists of a reconstructing (top) architecture and a transferring (bottom) architecture. In the top reconstructing flow, an encoder $E$ embeds a mesh input pose code $z_e$ and shape code $z_i$ which will be decoded via a generator $G$ to reconstruct the original mesh. In the bottom transferring flow, the pose code $z_e$ of the target mesh and source mesh will be swapped to generate pose-transferred meshes $G(z_e^1,z_i^2)$. The generated and reconstructed meshes will be fed into global discriminator $D$ for realistic effects. A Laplacian co-occurrence discriminator $D_{Lap}$ will serve as the extrinsic preservation. At last, the regional geodesic loss between shape mesh and generated mesh will be calculated to achieve intrinsic preservation.}
    \label{fig:overview}
        \vspace{-0.2cm}
\end{figure*}

In this paper, we propose a novel Intrinsic-Extrinsic Preserved Generative Adversarial Network (IEP-GAN) (see Fig.~\ref{fig:overview}). Two-branch discriminators are introduced to liberate the learning from the needs of data constraints. A global-branch discriminator is introduced to substitute the ground-truth meshes by enforcing generated meshes to converge to realistic ones. Besides, an extrinsic-branch discriminator is incorporated to enhance the pose style learning through the co-occurrence statics of the Laplacians of the meshes. Furthermore, a geodesic-adaptive sampling strategy is introduced to compute a regional geometric-preserved loss instead of the global geodesic prior, which can form an effective geometric regulation as intrinsic preservation while avoiding the heavy computations. 

To summarize, the novelties can be listed as below:
\begin{itemize}[leftmargin=*]
\item To the best of our knowledge, the proposed work is the first that can achieve unsupervised 3D pose transfer without any human supervision and the intrinsic-extrinsic preserved generative adversarial network is the first GAN-based framework for 3D human pose learning.
\item The IEP-GAN consists of newly proposed two-branch discriminators. The global branch enhances the generative capacity of the model and substitutes the need for the manual encoding of the training data. The extrinsic branch utilizes the Laplacian to strengthen extrinsic geometric learning in a co-occurrence static way.
\item At last, a regional geometric-preserved loss is proposed to preserve the regional intrinsic priors via a geodesic-adaptive sampling strategy, which guarantees both an effective geometric regulation and substantial improvements in the computational efficiency. 

\item Experimental results on four different datasets, i.e., DFAUST~\cite{DFAUST}, FAUST~\cite{FAUST}, ANIMAL~\cite{animal}, and MANO~\cite{MANO}, show that the proposed method achieves the new state-of-the-art performances with satisfying visual qualities. Moreover, we further represent the possibility of using the resulting embedding space for the 3D human mesh manipulation, such as the smooth pose transfer, interpolation and swapping of the given shapes.

\end{itemize}
\section{Related Work}

\noindent\textbf{Disentanglement of Pose and Style Generative Learning.} In 2D image processing, deep generative models have already achieved tempting performances of unsupervised learning and manipulating the data distribution with independently controllable factors~\cite{swap, mixn,stylegan2}. The key idea of those methods to achieve unsupervised learning is to introduce distinct discriminators that served as training manipulators to give modulating information with desired various factors. Our work is inspired by~\cite{swap, mixn} but intensively extended from 2D images to 3D meshes/points. To the best of our knowledge, this is the first work to implement GAN-based frameworks for the 3D human mesh manipulation task. The idea of swapping pose codes for the style transfer can also be found in~\cite{LIMP, Unsupervised,3dswap}, but we additionally introduce the double-branch discriminators that provide a strong regulation of geometric learning both globally and locally in a complete unsupervised setting. 

\noindent\textbf{Deep Geometric Representation.} Unlike 2D image understanding that pixels are placed in regular grids and can be processed with classical convolution, 3D points/meshes are unordered and scatted in 3D space. A variety of research efforts~\cite{pointnet,pointnet++,point2sequence,pointconv} have arisen in response to this challenge, including 3D voxelization-based methods~\cite{volix}, continuous convolutions~\cite{continoues} and the most popular one, graph-based models~\cite{grapghpoint} such as the PointNet family~\cite{pointnet, pointnet++}. Those models indeed achieved substantial improvements over previous works. However, they merely considered unstructured point-wise aggregation to the largest activation, the crucial structured geometric information of 3D points/meshes was not fully considered. To this end, recently, different geometry-preservation metrics were adopted to enhance the deep geometric representation learning, such as As Rigid As Possible~\cite{arap}, geodesics in heat~\cite{GIH}. But resolving analytical solutions of the global geodesic metrics in the back-propagation involves enormous computational consumption. In the work of~\cite{LIMP}, a simple down-sampling to the original meshes was conducted to reduce the computational complexity, which equitably leads to geodesic distortions. In this work, we focus on preserving of geometric properties for deformable shapes and guarantee an efficient computational capacity.

\noindent \textbf{Unsupervised Learning for 3D Deformation Transfer.} The pioneering work for mesh deformation transfer can go back to~\cite{animal, 3dcode}. Those methods rely on specifying typically a large number of point-wise correspondences between source and target shapes with local deformation gradients. Recently, deep generative models have been a main-stream solution for the task of 3D mesh deformation transfer due to their powerful learning and generalization ability. \cite{disentanglementaaai} and~\cite{Unsupervised} separately proposed an auto-encoder/a cycle-consistent adversarial network that can achieve automatic deformation transfer. But for each mapping domain, a new model needs to be trained. \cite{NPT} can achieve the training on multiple shapes simultaneously, but it has a strong prerequisite of large-scale synthesized targeted pose meshes and fails in real-world datasets from time to time. Inspired by the generative models~\cite{pointvae} on point-cloud processing, some unsupervised methods like~\cite{Unsupervised, LIMP} tried to use the geometric preservation priors to disentangle the shape and pose factors and conduct the pose transfer via latent code swapping. However, their whole theory and implementations reply on a training data assumption: the same subject in different poses should be given. Thus, the training of those models was all intervened via hand-crafted schemes of the input data types to ensure effective disentanglement. If a dataset cannot satisfy this assumption, the unsupervised setting in~\cite{Unsupervised, LIMP} will no longer be valid anymore. Different from all the existing works, we address a fully unsupervised setting that no prior information between training meshes needs to be provided and training can be achieved with arbitrating given meshes.
\section{Method}
This section first presents a general introduction of the whole IEP-GAN framework. Then, each component of IEP-GAN and their functions will be indicated. At last, designing details of the network will be presented.

\begin{figure*}[!h] \small
    \centering
    \includegraphics[width=0.8\linewidth]{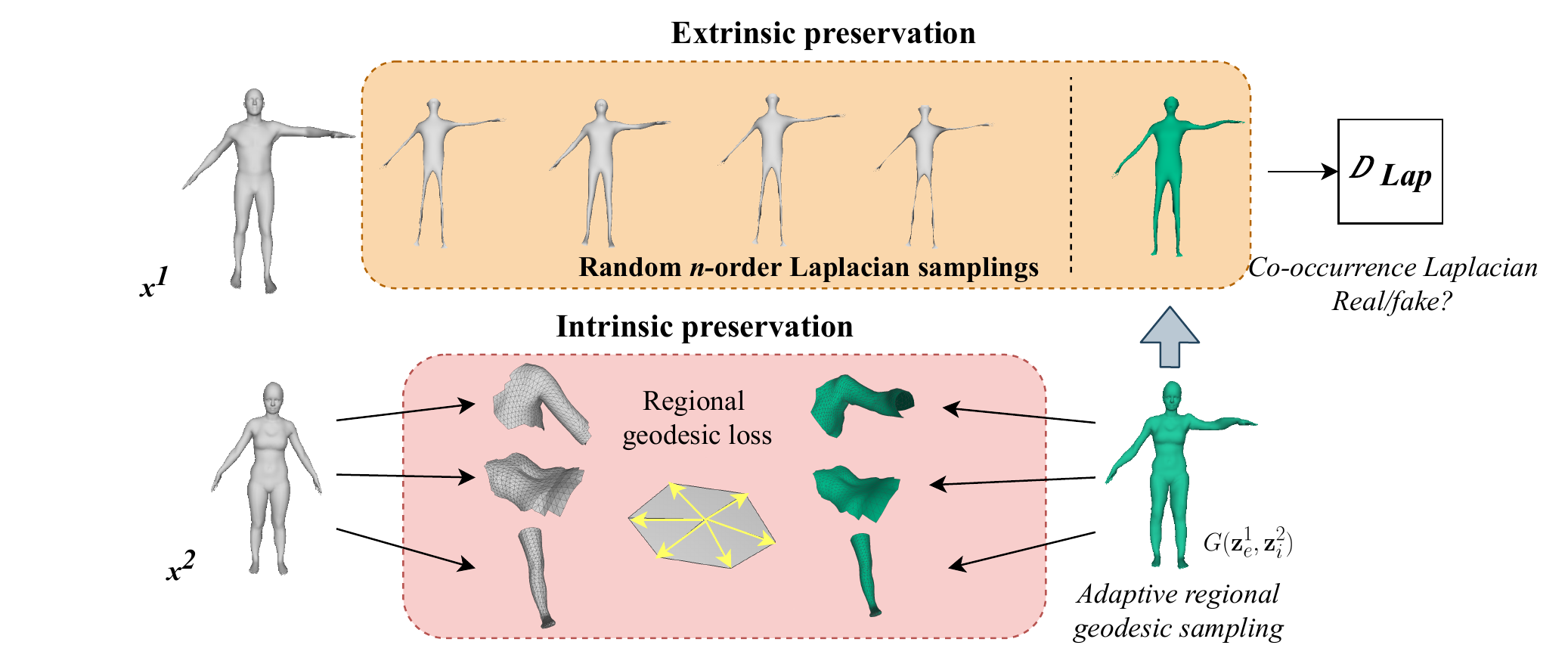}
    \caption{Intrinsic and extrinsic preservation in our IEP-GAN. In the top extrinsic preservation, a co-occurrence discriminator is used to distinguish the Laplacians between the real meshes and generated meshes. In this way, the generated mesh is enforced to have the same extrinsic/pose style as the real meshes. In the bottom intrinsic preservation, the real mesh and generated mesh will be adaptively sampled into several topology regions. The geodesic preservation metrics of each region of the generated mesh will be calculated and compared with the corresponding regions of the real meshes.}
    \label{fig:preservation}
    \vspace{-0.4cm}
\end{figure*}

\subsection{Overview}
We achieve the goal of unsupervised representation learning with a generative adversarial network, as shown in Fig.~\ref{fig:overview} based on four core tasks, 1) including accurately mesh reconstructing, 2) realistically pose transferring, 3) intrinsic-extrinsic preservation, and 4) disentanglement. 

To address these four tasks, the full objective function of the whole framework training is derived in:
\begin{equation}
\raggedright
\label{overall loss}
\begin{aligned}
\mathcal{L}_{full} & =   \mathcal{L}_{rec}+\mathcal{L}_{GAN,rec}+
\mathcal{L}_{GAN,transfer}\\
& + \mathcal{L}_{intrinsic} + \mathcal{L}_{extrinsic},
\end{aligned}
\end{equation}
where the first and second term $\mathcal{L}_{rec}$ and $\mathcal{L}_{GAN,rec}$ are a pair of adversarial loss, as in a classical GAN for accurately mesh reconstructing; the third one $\mathcal{L}_{GAN, transfer}$ is a discriminator loss for generating realistic pose-transferred meshes, and the last two are introduced to ensure the disentanglement of shape and pose information.

\subsection{Point-Wise Reconstruction}
The design of IEP-GAN starts from defining a classic autoencoder \cite{ae}. It is used to encode pose (i.e., extrinsic) and shape (i.e., intrinsic) information of a given mesh $\mathbf{x}{\small \sim} \mathbf{X}$ into corresponding latent representations $\mathbf{z}_e,\mathbf{z}_i {\small \sim} \mathbf{Z}$ with an encoder $E$, and reconstruct the original mesh via a generator $G$ using an point-wise reconstruction loss:
\begin{align}\label{rec loss}
&\mathcal{L}_{rec}(E,G)=\mathbb{E}_{\mathbf{x^1}{\small \sim}\mathbf{X}}[\|\mathbf{x^1}-G(E(\mathbf{x^1}))\|_2^2],
\end{align}
where the loss aims to regress the vertices close to its correct position with the supervision of point-wise L2 distance between the reconstructed mesh and the original mesh.

However, the simple point-wise regression will not guarantee a smooth surface. To avoid noticeable artifacts, we enforce the mesh to be realistic by introducing a discriminator $D$. The non-saturating adversarial loss~\cite{gan} for the generator $G$ and encoder $E$ is calculated as:
\begin{flalign}\label{recGANloss}
&\mathcal{L}_{GAN,rec}(E,G,D)=\mathbb{E}_{\mathbf{x}{\small \sim}\mathbf{X}}[-log(D(G(E(\mathbf{x}))))],
\end{flalign}

\subsection{Unsupervised Pose Transferring}
Based on the above autoencoder, the 3D pose transfer between two meshes $\mathbf{x}^1$ and $\mathbf{x}^2$ is realized by swapping their latent pose codes $\mathbf{z}^1_e$ and $\mathbf{z}^2_e$, as shown on the bottom part in Fig.~\ref{fig:overview}. In previous works, the mandatory constraints on the training data that different shapes in the same pose or different poses in the same shape should be held, which is imperative in supervising the model to correctly disentangle pose and shape from data. However, in real-world scenarios, such prerequisites can not always be satisfied. Targeting this issue, we advance the task to an unsupervised setting that learns the pose transfer between any arbitrarily given 3D human meshes. We achieve this by introducing a discriminator using the GAN loss:
\begin{equation}
\label{Disloss}
\begin{aligned}
&\mathcal{L}_{GAN,transfer}(E,G,D)=\\
&\mathbb{E}_{\mathbf{x}^1,\mathbf{x}^2{\small \sim}\mathbf{X}, \mathbf{x}^1\neq \mathbf{x}^2}[-log(D(G(\mathbf{z}^1_e, \mathbf{z}^2_i)))],
\end{aligned}
\end{equation}
where $\mathbf{z}^1_e$ and $\mathbf{z}^2_i$ are obtained from the two meshes $\mathbf{x}^1$ and $\mathbf{x}^2$ with the encoder $E$. They are later used to generate a new mesh while preserving the extrinsic/pose information from mesh $\mathbf{x}^1$ and the intrinsic/shape information from mesh $\mathbf{x}^2$. At last, the discriminator can serve as a substitute ground truth to enforce the generated meshes to converge to realistic ones. 

In an ideal case, the training of the IEP-GAN will converge to a status with several desirable properties. 1) The encoding function $E$ and generating function $G$ are optimized toward a smooth mapping between different meshes and their corresponding latent codes. 2) The latent codes should be successfully disentangled in relation to the intrinsic and extrinsic factors: the intrinsic code captures the shape distribution, while the pose code captures structural information of the meshes. 3) the latent pose code of generated mesh can be factored and it's possible to adjust the pose of generated meshes by manipulations on the distribution of latent pose codes. 

For the above first property, it can be easily realized with the reconstruction loss. However, ensuring the latter two properties is an extremely challenging task. It's tricky to disentangling the latent codes $\mathbf{z}_i$ and $\mathbf{z}_e$ and guarantees that they exactly represent the shape and pose in an unsupervised setting. Different constraints are put on the training datasets to achieve satisfying disentanglement of these two factors, for instance, fixing one factor (pose/shape) unchanged and training the model to preserve one factor at one time. Unlike any existing work, we tackle the issue in an unconstrained manner by designing two novel constraints to separately capture the intrinsic and extrinsic information, which will be introduced in the coming two sections.

\begin{table*}[] \small
\caption{Comparison with state-of-the-art methods on the FAUST and DFAUST datasets. Interpolation error measures the intrinsic preservation ability. Disentanglement error measures the extrinsic preservation ability. Results with the best performance are in bold font. Results of our method are marked with underline. Note that our IEP-GAN is trained without the need of data constraints which is used in the compared methods.}
\centering
	\resizebox{1\linewidth}{!}{%
\begin{tabular}{@{}ccccccccc@{}}
\toprule
 & \multicolumn{4}{c}{Interpolation Error} & \multicolumn{4}{c}{Disentanglement Error} \\\midrule
Method & VAE~\cite{pointvae} & LIMP Euc~\cite{LIMP} & LIMP Geo~\cite{LIMP} & IEP-GAN (Ours) & VAE~\cite{pointvae} & LIMP Euc~\cite{LIMP} & LIMP Geo~\cite{LIMP} & IEP-GAN (Ours) \\ \midrule
FAUST & 3.89e-2 & 5.08e-3 & \textbf{3.82e-3} & {\ul 4.02e-3} & 7.16 & 4.04 & 3.48 & {\ul \textbf{0.19}} \\
DFAUST & 9.82e-2 & 3.43e-3 & \textbf{2.89e-4} & {\ul 3.16e-4} & 6.15 & 4.90 & 4.11 & {\ul \textbf{0.34}} \\ \bottomrule
\end{tabular}}
\label{Tab:quantitive}
\end{table*}

\subsection{Regional Geodesic Preservation} To preserve the shape information during the pose transfer, we utilize the isometric relationship between original mesh $\mathbf{x}^2$ and generated mesh $G(\mathbf{z}^1_e, \mathbf{z}^2_i)$ with the help of geodesic priors as shown in Fig.~\ref{fig:preservation}, bottom part. Specifically, we call a pair of meshes are isometric if they share the same shape that the geodesic distance prior to these two meshes should be completely identical. Thus, we introduce the idea of the geodesic distance distortion~\cite{GIH} in the backpropagation to make the shape of the generated mesh converge to the original one, which can be defined as below:
\begin{equation}
\label{gih loss}
\begin{aligned}
&\mathcal{L}_{intrinsic}(E,G)=\\
&\mathbb{E}_{\mathbf{x}{\small \sim}\mathbf{X}}[\|D_{geodesic}(\mathbf{x}^2)-D_{geodesic}(G(\mathbf{z}^1_e, \mathbf{z}^2_i))\|_2^2],
\end{aligned}
\end{equation}
where $D_{geodesic}$ defines the pair-wise geodesic distances between all the vertices on a mesh topology and the intrinsic loss $\mathcal{L}_{intrinsic}$ measures the L2 norm of the two terms. Classical methods \cite{GIH,fastgeodesic}, however, have to travel the whole mesh topology to measure the geodesic distances, which involves heavy computations and cannot be extended to large-scale training. Down-sampling the vertex number from 6,890 to 2,100 was conducted in the work of~\cite{LIMP} to reduce the computational load while this operation will evidently damage the geometric details as well as the high-fidelity capability of the model. Thus, we introduce an adaptive geodesic measurement as $D_{ag}$ to compute on several mesh regions instead of the whole topology distance. The formulation of $D_{ag}$ is defined as below:
\begin{equation}
\label{agloss loss}
\begin{aligned}
& D_{ag}(\mathbf{x})=\sum_{i}^{N}D_{geodesic}(\mathbf{x}_i), \mathbf{x}_i \in \mathcal{N}_k(\mathbf{p}_i),
\end{aligned}
\end{equation}
where $N$ is the sub-region number for the geodesic measurement, $\mathcal{N}_k(\mathbf{p}_i))$ denotes the sub-mesh formed from the $k$ neighbor vertices of vertex $\mathbf{p}_i$. The vertex $\mathbf{p}_i$ is obtained by an geometric adaptive sampling strategy based on a local geometric distortion score ranking $\mathcal{R}$:
\begin{equation}
\label{adaptiverank}
\begin{aligned}
\mathcal{R} = \{\mathbf{r}_1, \mathbf{r}_2, ... ,\mathbf{r}_j,..., \mathbf{r}_V |\: \mathbf{r}_j = \sum_{\mathbf{u}\in \mathcal{M}(\mathbf{p}_j)}\left \|\mathbf{p}_j - \mathbf{u} \right \|^{2}_{2}\},
\end{aligned}
\end{equation}
where $\mathcal{M}(\mathbf{p}_j)$ denotes the one-ring neighbor vertices of vertex $\mathbf{p}_j$ and $V$ is the total vertex number of a topology mesh. The local geometric distortion score ranking $\mathcal{R}$ is a list that measures the local geometric distortion of each vertex on the mesh. We sample the top $N$ vertex and calculate their sub-regions. The sampling of the sub-region is updated based on the local geometric distortion score ranking $\mathcal{R}$ at each training iteration which ensures a global traversing on the whole topology. In practice, the sub-region sampling number $N$ is 4 and the sampling vertices of each sub-region is 300, which is a balance between the on-the-fly computational load and geodesic preservation capability of the model. In this way, the computations of calculating the geodesic priors is substantially reduced. For instance, given a typical body mesh with 6,890 vertices, the computations of $D_{ag}$ (with 4 sub-regions of 300 sampling vertices) will be conducted on 2$\times$300$\times$300 vertices, which is around 262 times fewer than the whole topology distance $D_{geodesic}$ one with 6,890$\times$6,890 vertices.

\subsection{Laplacian Extrinsic Preservation} 
By implicitly swapping the latent pose codes and preserve the geodesic priors, the pose transfer can be realized to some extent. But degeneracy may occur in the network. In practice, the network sometimes lets generated meshes inherit unexpected pose attributes of $\mathbf{x}^2$ (shape source) through the shape code. In this shortcut way, the model can directly generate meshes with similar poses to $\mathbf{x}^2$ to confuse the discriminator, instead of meshes with the target pose. This means unwanted pose information from the $\mathbf{x}^2$ (shape source) flows into the shape code and a constraint for extrinsic preservation is desirable.

To this end, we propose a Laplacian co-occurrence discriminator $D_{Lap}$, as shown at the bottom of Fig.~\ref{fig:preservation}. It aims to distinguish generated mesh $G(\mathbf{z}^1_e, \mathbf{z}^2_i))$ from a variety of Laplacians of input mesh $\mathbf{x}^1$ so that the generator $G$ is encouraged to capture the extrinsic information:
\begin{equation}
\label{extrinsic loss}
\begin{aligned}
&\mathcal{L}_{extrinsic}(E,G,D_{Lap})=\\
&\mathbb{E}_{\mathbf{x}^1{\small \sim}\mathbf{X}}[-log(D_{Lap}(L_{group}(G(\mathbf{z}^1_e, \mathbf{z}^2_i))),L_{group}(\mathbf{x}^1))],
\end{aligned}
\end{equation}
where $L$ generates a random Laplacian with iteration number between 1 to 100 of the given mesh (and $L_{group}$ is a collection of multiple Lapacians). Our formulation is inspired by the work in the 2D texture perception \cite{percieption,texturepecp}, but radically transferred to 3D task with the hypothesis that meshes with the same pose/extrinsic style will hold perceptually similar on the Laplacian statistics. The co-occurrence discriminator serves to enforce the extrinsic information to be consistently transferred by using the Laplacian to detach all the detailed shape-related information. Similar ideas for modeling co-occurrences have been used \cite{texturegan, swap}, but it's the first time that the idea is applied to 3D mesh fields.

\subsection{Network Architectures} 
The encoder $E$ used for processing the input meshes is the classical PointNet \cite{pointnet}. Besides, each convolutional layer is stacked by InstanceNorm layers instead of a batch normalization operator for preserving the instance features which is critical for the style transfer task. The pose code is fixed as 512 dimensions and the shape code is the raw input shape mesh for the pose transfer task and 2048 dimensions for the disentanglement. For the generator, we fuse the architecture of \cite{swap} and \cite{NPT}. In particular, we adopt the GAN framework and network structure of \cite{swap}, and change its Residual blocks into the SPAdaIN ResBlock introduced in \cite{NPT} which is specifically designed for 3D style transfer task. The discriminator structure is a symmetry of the generator except that the last two layers of the discriminator become fully-connected layers to produce the prediction of real or fake meshes. The general structure of the Laplacian co-occurrence discriminator is identical to \cite{swap} while we replace its encoder with ours mentioned above for 3D meshes. Each Laplacian will be down-sampled via the Poisson Disk method~\cite{sampling}. Please see Supplementary Materials for a detailed specification of the whole architecture, as well as details of the hyper-parameter settings.

\section{Experiments}
In this section, comprehensive experiments are conducted to evaluate our proposed approach by comparing other state-of-the-art models on a variety of datasets, including human mesh datasets FAUST~\cite{FAUST} and DFAUST~\cite{DFAUST}, animal dataset ANIMAL~\cite{animal} and hand mesh dataset MANO~\cite{MANO}. Firstly, quantitative evaluations of our IEP-GAN on the two datasets are represented. One step further, we qualitatively visualize the strong generalization ability and disentanglement effect of the IEP-GAN on the four datasets. Lastly, we also perform ablation studies to evaluate the effectiveness of intrinsic and extrinsic preservation.
\begin{figure}[!h] 
    \centering
    \includegraphics[width=0.85\linewidth]{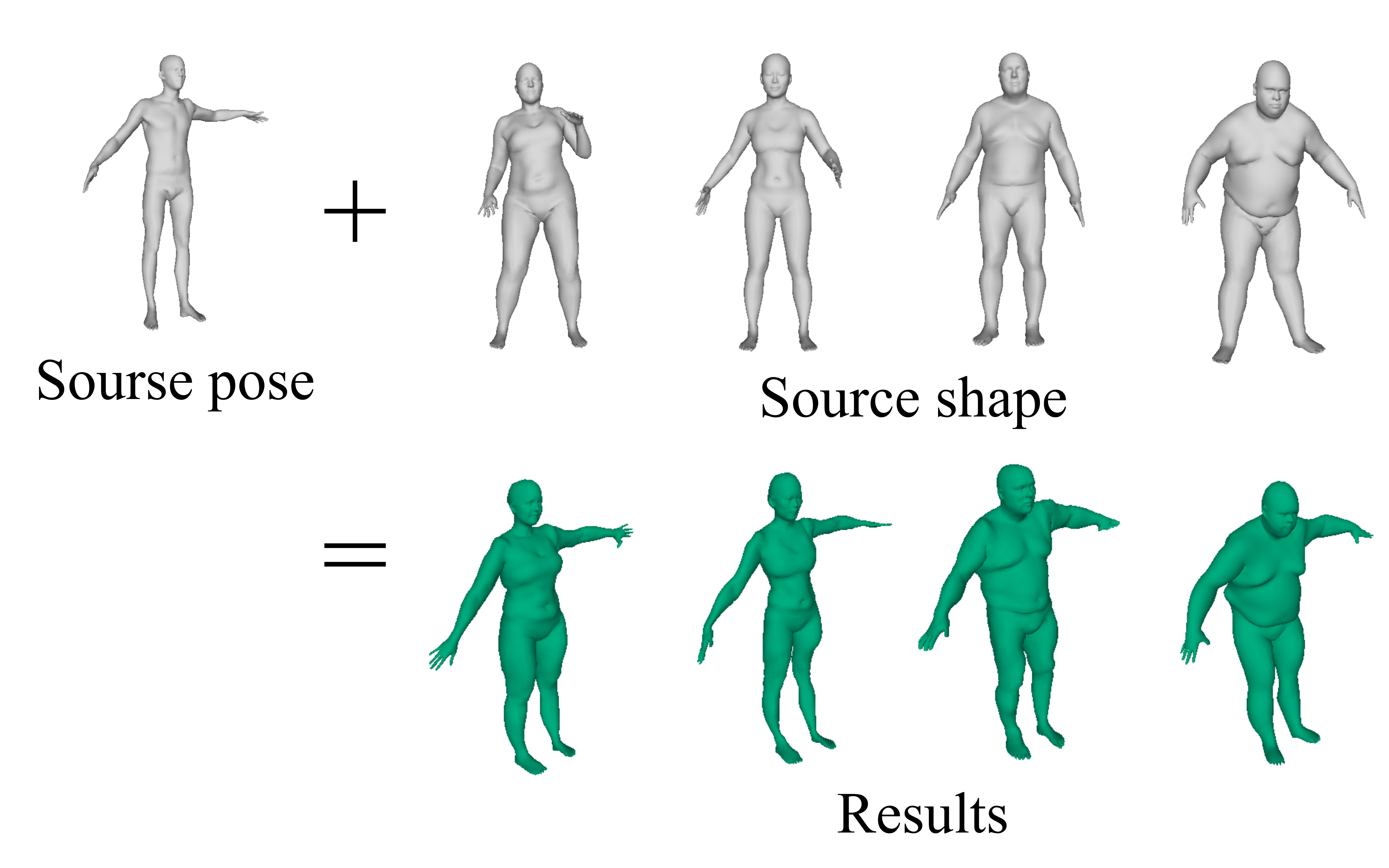}
    \caption{Pose transferring effects of our IEP-GAN on DFAUST dataset. The top part is a list of source shapes. The bottom part shows the corresponding results under the same pose. No constraint on the training dataset is needed for our method, which implements an unsupervised setting.}
    \label{fig:header}
    \vspace{-0.45cm}
\end{figure}

\subsection{Datasets}

\noindent \textbf{FAUST} \cite{FAUST} dataset is a 3D human body scan dataset collected from 10 different human subjects, each performing in 10 different poses. The mesh structure of FAUST registrations fits the SMPL body model that has 6,890 vertices.

\noindent \textbf{DFAUST} \cite{DFAUST} dataset is a large human motion sequence dataset that captures the 4D motion of 10 human subjects performing 14 different body motions, such as ``punching'', ``one leg loose'', ``one leg jump'' and etc. The length of each motion spans hundreds of frames. Following the same protocol~\cite{LIMP}, four representative frames are used for one training instance (per subject per motion).

\noindent \textbf{MANO} \cite{MANO} dataset is the 3D hand dataset obtained by hand scans with registrations fitting \cite{MANO} mesh model. The scans in MANO dataset include both right and left hands and the missing data is compensated by mirroring the corresponding hands. Following the same processing strategy in \cite{Unsupervised}, the model was trained on right hands and generalized to left hands by mirror flipping.

\noindent \textbf{ANIMAL} \cite{animal} is an synthesized animal mesh dataset from ~\cite{animal}. The synthesized animal meshes are parametrically deformed to desired poses from 3D quadrupedal animal models as ground truths.

\begin{table*}[t] \small
\centering
\caption{Ablation study of the adaptive regional geodesic preservation on FAUST dataset. To make it fair, runtime is measured on the same platform (Pytorch 1.6.0) and hardware setting (a single GPU NVIDIA GTX 1080Ti, CPU Intel Core i7). The vertex number for the GIH calculating is fixed as 6,890. The interpolation errors below are all obtained with IEP-GAN.}
\begin{tabular}{@{}cccccc@{}}
\toprule
Method & Region& Sampling & Interpolation Error & \begin{tabular}[c]{@{}c@{}}Computational \\ Complexity\end{tabular} & Runtime \\ \midrule
No geodesic preservation & - & - & {\ul 3.11e-2} & - & 0.00 s \\ \midrule
\multirow{3}{*}{Geodesic preservation} & Global & - & {\ul 3.94e-3} & $O(n^2)$ & 8.31 s \\
 & Regional & Random & {\ul 7.29e-3} & $O(n \, log \, n)$ & 1.81 s \\
 & \multicolumn{1}{l}{Regional} & Adaptive & {\ul 4.02e-3} & $O(n \, log\,  n)$ & 2.07 s \\ \bottomrule
\end{tabular}
\vspace{-0.4cm}
\label{Tab:regionalpreservation}
\end{table*}

\begin{figure*}[!h] 
    \centering
    \includegraphics[width=0.9\linewidth]{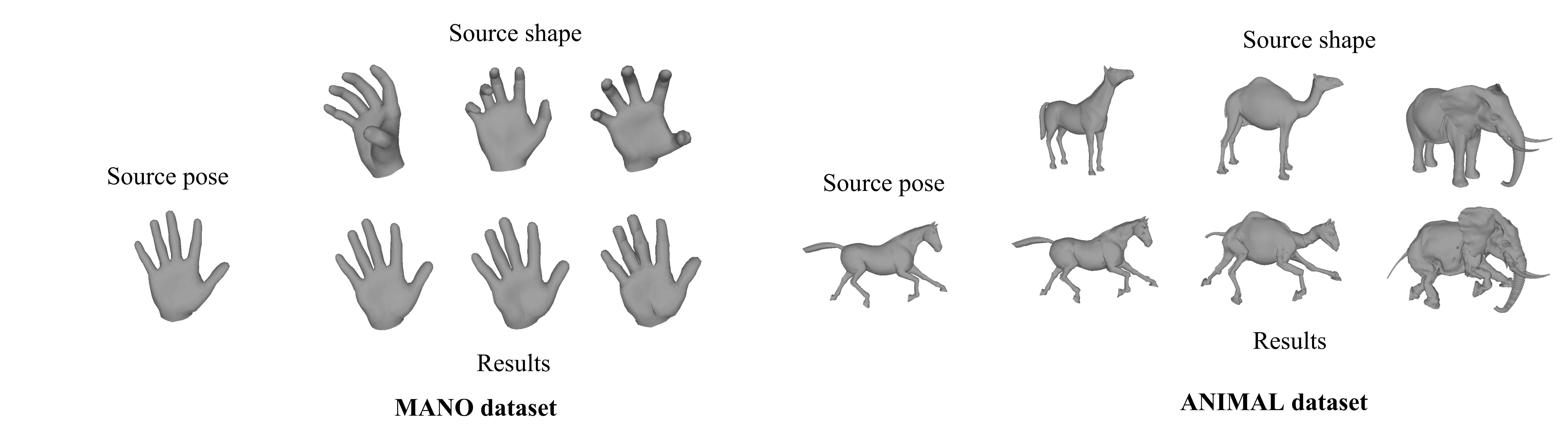}
    \caption{Pose transfer results of our methods. Right one is the MANO dataset with human hand meshes. The right one is the ANIMAL dataset that contains various animals with different poses. The top line shows the source mesh with distinct shapes, the bottom line shows the meshes with transferred poses.}
    \label{fig:transfer}
    \vspace{-0.2cm}
\end{figure*}
\subsection{Implementation Details}
Our algorithm is implemented in PyTorch~\cite{paszke2019pytorch}. All the experiments are carried out on a PC with a single NVIDIA Tesla V100, 32GB. We train our networks for $4\times 10^4$ epochs with a learning rate of 0.00005 and Adam optimizer~\cite{adam}. The batch size is fixed as 4 for all settings. The whole training time takes around 30 hours. Please refer to the supplementary material for more details.

Training GAN is extremely hard and easily encounters the generator collapses issue, thus we design the training of the IEP-GAN with three stages. For the first $2\times10^4$ iterations, only reconstruction is conducted to stabilize the GAN and avoid local minima, where the pair of reconstruction adversarial loss is used with raw SMPL meshes. From iterations $2\times10^4$ to $3\times10^4$, the pose transfer learning starts with the swapping loss and extrinsic preservation loss added. The intrinsic preservation loss will be added after $3\times10^4$ iterations.

\subsection{Quantitative Evaluation for Pose Transfer}
In the following, we quantitatively present the performance of our IEP-GAN on the pose transfer task on the FAUST and DFAUST datasets, including extrinsic preservation (pose-transfer) ability and intrinsic preservation (shape-preserve) capacity.

\noindent \textbf{Intrinsic Preservation}. We first validate the intrinsic preservation ability of the IEP-GAN. To do this, we utilize the average (overall surface points) geodesic distortion of the pose-transferred shapes as metrics. The average geodesic distortion is defined in \cite{GIH} which measures the geodesic distance inconsistencies between the given meshes. It can be regarded as an approximated ground truth of the shape information which is commonly used in previous works \cite{LIMP,fastgeodesic}. It's also named as interpolation error for measuring the intrinsic preservation ability during the meshes interpolation procedure. 

We compare our method to state-of-the-art methods, including pointVAE~\cite{pointvae}, Euclidean-based VAE (LIMP Euc) \cite{LIMP} and geodesic-based VAE (LIMP Geo) \cite{LIMP} on the FAUST \cite{FAUST} and DFAUST \cite{DFAUST} datasets. The intrinsic preservation performance is shown in the right part (interpolation error) in Table \ref{Tab:quantitive}. It is important to note that the compared methods \cite{LIMP} used the data constraint that can strictly supervise the disentanglement shape and pose which is kept untouched for our IEP-GAN. The interpolation error of our unsupervised model is only 2.00e-4 and 2.70e-4 inferior to the compared baseline method (supervised with data constraints) on FAUST and DFAUST dataset, showing a competitive intrinsic preservation ability.

\noindent \textbf{Extrinsic Preservation}. As there is no ground truth for pose transfer (extrinsic preservation) task that different subjects performing exactly the same pose in the existing datasets, following the protocol in \cite{LIMP}, we utilize the \emph{disentanglement error} for evaluating extrinsic preservation (pose-transfer) performances. Specifically, the notion of disentanglement error is defined as follows. Given two meshes for the pose transferring in the same intrinsic property, we conduct the pose transfer between them and then measure the average point-to-point distance between the pose transferred meshes. Thus the ground truths can be obtained from the corresponding shapes in the dataset.

The left part (disentanglement error) in Table \ref{Tab:quantitive} presents the extrinsic preservation performances of different models. It can be observed that our IEP-GAN has a substantially better extrinsic preservation performance than the compared methods (0.19 vs. 3.48 on the FAUST dataset and 0.34 vs. 4.11). It proves that our extrinsic preservation constraint (the Laplacian co-occurrence discriminator) is considerably powerful for capturing the extrinsic information (pose style) and disentangling pose from shape.

\begin{figure}[!t] \small
    \centering
    \includegraphics[width=0.8\linewidth]{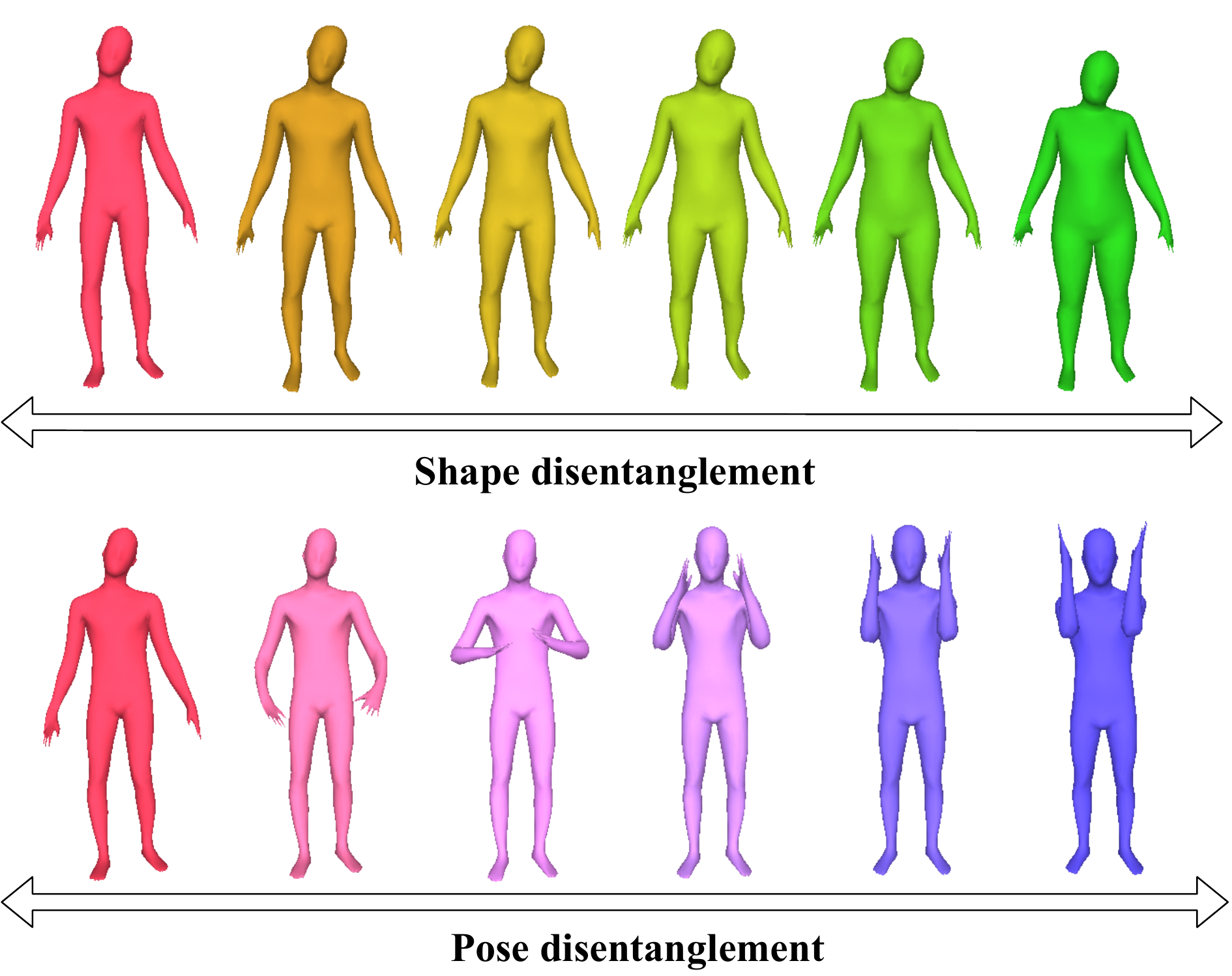}
    \caption{Disentanglement effects of our IEP-GAN. The top part shows the shape disentanglement effect by interpolating among two latent shape codes. The bottom part shows the disentanglement transfer effect by interpolating among two latent pose codes.}
    \label{fig:interpolation}
    \vspace{-0.4cm}
\end{figure}

\subsection{Qualitative Evaluation}

We qualitatively visualize our method's generalization ability on the four datasets, i.e., the above FAUST and DFAUST datasets, as well as two extra datasets, MANO human hand dataset~\cite{MANO} and ANIMAL dataset~\cite{animal}. Fig.~\ref{fig:header} presents the pose transfer results on the DFAUST dataset. Based on the FAUST dataset, we provide the distribution of latent code space in Fig.~\ref{fig:latent} by utilizing the t-distributed stochastic neighbor embedding (t-SNE) method~\cite{tsne} for visualizing high-dimensional data. Specifically, the latent code produced by an encoder trained on the FAUST dataset will be embedded with t-SNE and project into low dimension space for visualization. Furthermore, Fig.~\ref{fig:interpolation} indicates the disentanglement effect of shape and pose with our IEP-GAN. At last, in Fig.~\ref{fig:transfer}, we present transfer results on an animal dataset ANIMAL and human hand mesh dataset MANO to show the generalization ability.

\subsection{Ablation Study}

\begin{table}[t] \small
\caption{Ablation study of the Laplacian extrinsic preservation on FAUST dataset. The reference Laplacian can largely enhance the extrinsic preservation learning.}
\centering
\resizebox{0.8\linewidth}{!}{%
\centering
\begin{tabular}{@{}ccc@{}}
\toprule
Method & \begin{tabular}[c]{@{}c@{}}Reference \\ Laplacian\end{tabular} & \begin{tabular}[c]{@{}c@{}}Disentanglement \\ Error\end{tabular} \\ \midrule
No intrinsic preservation & - & 2.6 \\ \midrule
\multirow{3}{*}{Intrinsic preservation} & 1 & 0.59 \\
 & 2 & 0.21 \\
 & 3 & 0.19 \\ \bottomrule
\end{tabular}
}
\vspace{-0.5cm}
\label{Tab:extrinsic}
\end{table}

Here we intensively conduct experiments to verify the effectiveness of our proposed key components.

\noindent \textbf{Effect of the Laplacian Co-Occurrence Discriminator}. We also validate the effect of the Laplacian co-occurrence discriminator, as shown in Table~\ref{Tab:extrinsic}. As expected, introducing the Laplacian co-occurrence discriminator results in significant gains (disentanglement error reduced from 2.6 to 0.19). This proves that, by measuring the Laplacian co-occurrence with reference meshes, the generated mesh could have better extrinsic preservation. However, increasing the number of reference Laplacians requires larger computational consumption and reaches the GPU memory limits, thus we adopt 3 reference Laplacians as the default setting in the experiments.

\noindent \textbf{Effect of Regional Geodesic Preservation}. We vary the settings of the proposed regional geodesic preservation to show its effect in Table~\ref{Tab:regionalpreservation}. In this first line, we observe that, without geodesic preservation, the local geometric distortion is severe with the interpolation error of 3.11e-2. This proves the necessity of geodesic-based priors. From the last three lines of the Table~\ref{Tab:regionalpreservation}, we can see that our regional geodesic preservation can exponentially reduce the computational complicity while conserving a satisfying geodesic prior (4.02e-3) compared to the global one (7.29e-3). Specifically, our proposed adaptive sampling strategy can further enhance the geodesic preservation compared to random sampling. Lastly, the runtime (2.07 vs.8.31 seconds) proves that our regional geodesic preservation can efficiently reduce the computations.
\section{Conclusion}
In this paper, we introduce IEP-GAN that achieves the pose transfer in an unsupervised manner with arbitrate given meshes. We show two key components to achieve the latent code disentanglement between pose and shape without extra data constraints, namely the intrinsic preservation and extrinsic preservation. The Laplacian co-occurrence discriminator is proved efficient in capturing the pose information and the adaptive regional geodesic loss can not only enhance the geometric prior learning but also substantially reduce the computations. At last, with the obtained latent codes, the task can be extended from 3D pose transfer to pose interpolation, shape swapping, and factor disentanglement. The regional geodesic preservation and Laplacian co-occurrence discriminator can be easily implemented to existing classical models. However, due to the challenging unconstrained setting in this work, artifacts and visible distortions in the pose transferring and interpolation still can be observed when the pose or the shape is too extreme. Improving the robustness with larger 3D space diversity and reducing the training time with alternative metrics of geometric distances would be the future work.

\noindent \textbf{Acknowledgement}.
This work was supported by the Academy of Finland for project MiGA (grant316765), ICT 2023 project (grant 328115), EU H2020 SPRING (No.871245) and EU H2020 AI4Media (No.951911) projects, the China Scholarship Council, and Infotech Oulu. As well, the authors wish to acknowledge CSC-IT Center for Science, Finland, for computational resources.

\clearpage
{\small
\bibliographystyle{ieee_fullname}
\bibliography{egbib}
}

\end{document}